\pdfoutput=1
\documentclass[runningheads]{llncs}

\usepackage{accv}
\usepackage{accvabbrv}

\usepackage{graphicx}
\usepackage{booktabs}
\usepackage[accsupp]{axessibility}
\usepackage{multirow}

\usepackage{hyperref}

\usepackage{orcidlink}

\usepackage{pifont}
\newcommand{\cmark}{\ding{51}} 

\usepackage{algorithm}
\usepackage{algpseudocode}

\newcommand{\algpoolref}{\bar{z}_{0|T}^{i,\mathrm{ref}} \gets \mathrm{Normalize}\bigl(\mathrm{AvgPool}(\hat{z}_{0|T}^{i,\mathrm{ref}}, w)\bigr)}
\newcommand{\algmedian}{\sigma \gets \mathrm{median}\bigl(\bigl\{\|\bar{z}_{0|T}^{i,\mathrm{ref}} - \bar{z}_{0|T}^{j,\mathrm{ref}}\|_2\bigr\}_{i<j}\bigr)}

\newcommand{\algpoolopt}{\bar{z}_{0|T}^{i} \gets \mathrm{Normalize}\bigl(\mathrm{AvgPool}(\hat{z}_{0|T}^{i}, w)\bigr)}
\newcommand{\alggram}{S_{ij} \gets \exp\bigl(-\|\bar{z}_{0|T}^i - \bar{z}_{0|T}^j\|^2 / 2\sigma^2\bigr)}
\newcommand{\algdiv}{\mathcal{L}_\mathrm{div} \gets -\log\det(S + \epsilon I)}
\newcommand{\algfid}{\mathcal{L}_\mathrm{fid} \gets \sum_{i=1}^{B} \|\hat{z}_{0|T}^i - \hat{z}_{0|T}^{i,\mathrm{ref}}\|^2}
\newcommand{\algloss}{\mathcal{L} \gets \mathcal{L}_\mathrm{div} + \lambda \cdot \mathcal{L}_\mathrm{fid}}
\newcommand{\algupdate}{z_T^i \gets z_T^i - \eta \nabla_{z_T^i}\mathcal{L}}

\newcommand{\grantsponsor}[1]{#1}
\newcommand{\grantnumber}[2]{$^{#1}$#2}

\usepackage{xcolor}
\usepackage[numbers]{natbib}

\usepackage{mathtools}

\usepackage[normalem]{ulem}
\newcommand{\rev}[1]{#1}
\newcommand{\del}[1]{}

\usepackage[table]{xcolor}
\definecolor{oursblue}{RGB}{218, 232, 252}

\begin{document}

\title{Beyond Pairwise Repulsion: Collective Diversity via Determinant Maximization in Diffusion Sampling}

\titlerunning{Beyond Pairwise Repulsion}

\author{Kwanyoung Lee\inst{1} \and
SeungJu Cha\inst{1} \and
Yebin Ahn\inst{1} \and\\
Sungho Koh\inst{1} \and
Dong-Jin Kim\inst{1}}

\authorrunning{K.~Lee et al.}

\institute{Hanyang University, South Korea\\
\email{\{mobled37,sju9020,yeshine,ksh000906,djdkim\}@hanyang.ac.kr}}

\maketitle

\begin{figure}
    \centering
    \includegraphics[width=1\linewidth]{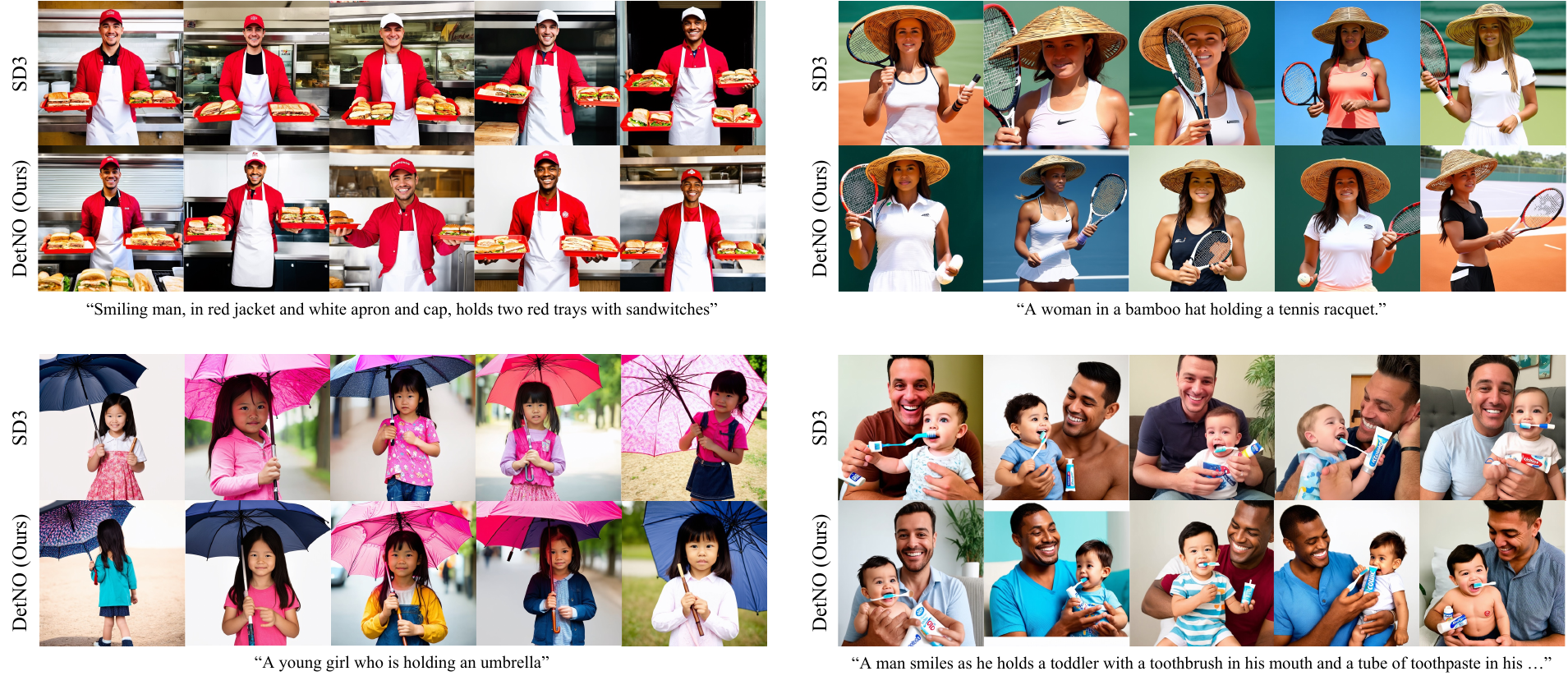}
    \caption{\textbf{Qualitative comparison on SD3.} SD3 baseline sampling collapses onto a narrow visual prototype for underspecified prompts, reproducing nearly identical appearance and scene layout across seeds\del{ ---}\rev{,} a direct consequence of CFG concentrating the conditional distribution onto its high-density peak. DetNO recovers the suppressed degrees of freedom in initial noise, producing semantically diverse outputs that vary in viewpoint, identity, clothes, and scene composition while remaining faithful to the prompt.}
    \label{fig:teaser}
\end{figure}

\begin{abstract}
Strong guidance in text-to-image (T2I) diffusion models concentrates sampling toward high-density modes of the conditional distribution, collapsing independently sampled noise vectors onto nearly identical visual prototypes.
This limits utility for tasks that demand distributional breadth, most critically synthetic data augmentation.
Diversity, however, is a collective property of the whole batch: a set can preserve every pairwise distance yet still collapse onto a low-dimensional subspace.
Existing sum-of-pairwise noise-optimization and repulsion objectives respond only weakly to this higher-order failure.
We propose \textbf{DetNO} (\textbf{Det}erminant \textbf{N}oise \textbf{O}ptimization), which formulates diverse generation as RKHS volume maximization on the initial noise, maximizing the log-determinant of an RBF Gram matrix over one-step Tweedie predictions.
This set-level functional equals the squared batch volume; its repulsion is strongest when the batch is most collapsed and self-attenuates as diversity is achieved.
An anchor-based fidelity term keeps the optimized noises near plausible regions, and Lagrangian relaxation gives a two-term objective with a single hyperparameter $\lambda$ governing the diversity--quality trade-off, without training or sampler modification.
Across three backbones (SD1.5, SDXL, SD3), DetNO achieves leading intra-batch diversity in VendiScore and Mean Self-Similarity, and a downstream few-shot classification experiment confirms that broader mode coverage brings statistically significant accuracy gains for synthetic data augmentation in the low-shot regime.
Fixing the pipeline and varying only the batch objective, both RBF-based aggregators outperform InfoNCE, and the log-determinant attains the \del{highest downstream accuracy}\rev{best downstream accuracy on both benchmarks}.
Its diversity advantage over \del{pairwise aggregation}\rev{InfoNCE} widens as the batch grows.
Code is available at \url{https://github.com/mobled37/DetNO}.
\keywords{Text-to-Image Diffusion Models \and Diverse Image Generation \and Initial Noise Optimization \and Log-Determinant Maximization \and Synthetic Data Augmentation}
\end{abstract}

\section{Introduction}
\label{sec:introduction}

Classifier-Free Guidance (CFG)~\citep{ho2022classifier} has become the standard mechanism for steering text-to-image (T2I) diffusion models~\citep{ho2020denoising, song2020score, rombach2022high} toward high-fidelity, prompt-aligned images.
Strong CFG, however, concentrates sampling toward the high-density peak of the conditional distribution, collapsing its effective support onto a narrow region of the image manifold (Fig.~\ref{fig:teaser}).
Under deterministic samplers such as DDIM~\citep{song2020denoising}, the initial noise fully determines the final image, yet independent noise draws collapse onto the same narrow visual mode.
Multiple images generated from the same prompt converge to nearly identical visual prototypes.
The collapse limits utility in applications that depend on distributional breadth, most critically synthetic data augmentation, where near-duplicate outputs provide diminishing returns for downstream model training~\citep{he2022synthetic}.

Recovering this diversity requires treating a batch as a whole, since diversity is a collective property of the set rather than a sum over its pairs.
A batch can preserve every pairwise distance yet still collapse onto a low-dimensional subspace, a higher-order failure mode to which independent pair contributions respond only weakly.
Existing approaches meet this criterion only partially.
Methods that intervene during denoising repel intermediate latents~\citep{corso2023particle, kirchhof2024shielded} or manipulate text embeddings~\citep{um2025minority}, paying a per-step cost that scales with trajectory length.
Contrastive Noise Optimization (CNO)~\citep{kim2025diverse} instead optimizes the initial noise with an InfoNCE loss in the Tweedie prediction space, intervening before denoising begins, where the noise largely determines semantic structure~\citep{guo2024initno, ahn2024noise}.
Despite their differing entry points, all build their diversity signal from sum-of-pairwise repulsion, whose gradient remains confined to the span of pairwise directions and can miss higher-order redundancy.

This motivates a single scalar measuring the collective volume a batch occupies, penalizing redundancy at every order up to $B$ simultaneously.
We realize this scalar as the log-determinant of an RBF Gram matrix over one-step Tweedie predictions, making all-order diversity differentiable.
We further show that this objective is self-regulating: repulsion amplifies when the batch is most redundant and relaxes as diversity is achieved, an adaptation that an unweighted sum-of-pairwise objective lacks in the same analysis\del{ (Proposition~1)}.

\begin{figure}[t!]
\centering
\includegraphics[width=0.8\linewidth]{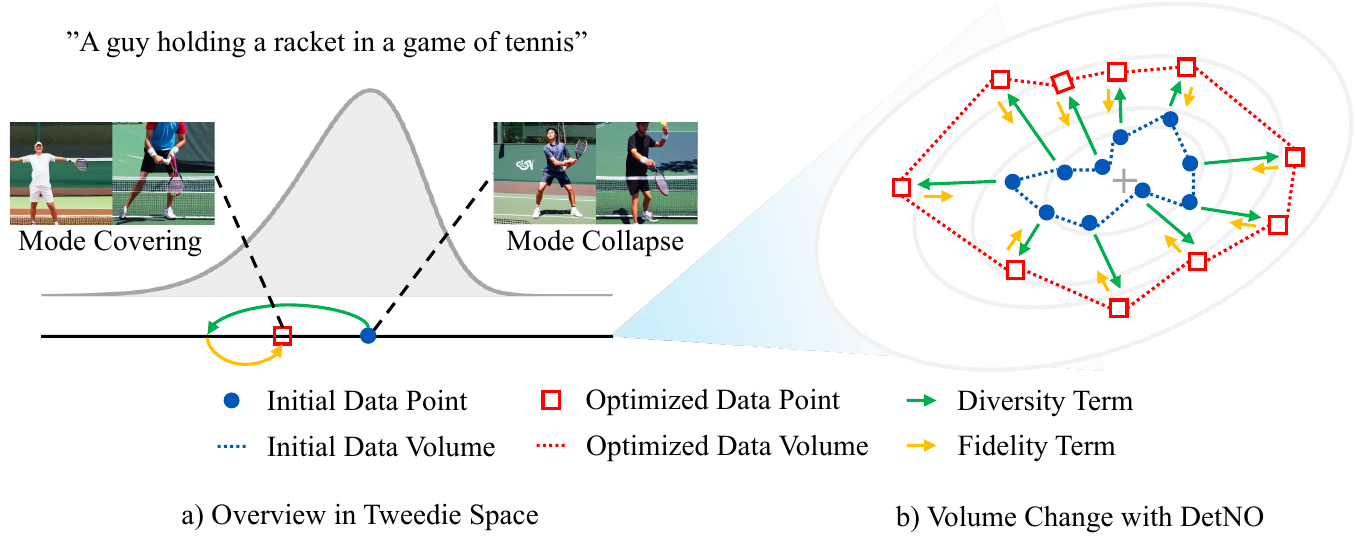}
\caption{\textbf{Conceptual illustration of DetNO.} (a) Strong CFG guidance concentrates sampling toward the high-density peak, leaving the degrees of freedom in the initial noise unexploited. DetNO recovers this freedom by optimizing the initial noise batch to reach low-density regions that standard sampling cannot access. (b) The diversity term (green) maximizes the RKHS volume collectively spanned by all Tweedie predictions, while the fidelity term (yellow) regularizes each sample toward its unoptimized reference, expanding distributional coverage while preserving prompt fidelity.}
\label{fig:conceptual}
\end{figure}

We propose \textbf{DetNO} (\textbf{Det}erminant \textbf{N}oise \textbf{O}ptimization), illustrated in Fig.~\ref{fig:conceptual}, which pairs this log-determinant diversity term with an anchor-based fidelity term via Lagrangian relaxation.
A single hyperparameter $\lambda$ controls the diversity-quality trade-off, with no training or sampler modification.
Empirically, the collective objective delivers on both diversity and downstream utility.
Across three backbones (SD1.5, SDXL, SD3), DetNO achieves leading intra-batch diversity in VendiScore and Mean Self-Similarity, with the advantage over \del{pairwise repulsion}\rev{CNO} widening as the batch size grows.
A controlled toy experiment shows faithful recovery of all modes where pairwise methods distort mode structure, and downstream few-shot classification produces \del{statistically significant }accuracy gains over both standard sampling and CNO\del{ in}\rev{, reaching statistical significance in} the low-shot regime ($p{<}0.01$).

Our main contributions are as follows:
\begin{itemize}
    \item We recast diverse T2I generation as a set-level rather than pairwise problem: constrained RKHS volume maximization on initial noise. The objective pairs a log-determinant diversity term with an anchor-based fidelity term, controlled by a single hyperparameter $\lambda$, with no sampler modification.
    \item We show that the log-determinant amplifies repulsion when the batch is most redundant and relaxes as diversity is achieved, a self-regulating property of the Gram matrix eigenspectrum that an unweighted sum-of-pairwise objective lacks in the same analysis\del{ (Proposition~1)}.
    \item We validate across three backbone-sampler configurations, demonstrating consistent diversity gains with a controllable quality-diversity trade-off and statistically significant downstream improvements in few-shot classification.
\end{itemize}
\section{Related Work}

\subsection{Recovering Mode Diversity in Diffusion Models.}
Prior methods recover the limited diversity of conditional diffusion models through repulsion, differing mainly in where they apply it but sharing a diversity signal that decomposes into a sum of pairwise interactions.
One line intervenes during denoising: SPELL~\citep{kirchhof2024shielded} adds sparse repellency terms to the reverse SDE, Particle Guidance (PG)~\citep{corso2023particle} repels intermediate latents via a joint-particle potential, and DiversityPrompt~\citep{um2025minority} optimizes a token embedding online to steer generation toward low-density regions.
All incur per-step overhead that scales with trajectory length.
Contrastive Noise Optimization (CNO)~\citep{kim2025diverse} instead shifts repulsion to the initial noise, optimizing latents via InfoNCE in the Tweedie prediction space before sampling begins.
Both, however, quantify diversity as a sum over sample pairs, so their repulsion never leaves the span of pairwise directions and \del{cannot detect}\rev{responds only weakly to} higher-order redundancy that collapses the batch onto a low-dimensional subspace.
A separate line targets rare modes of the \emph{unconditional} or class-conditional distribution via classifier guidance~\citep{sehwag2022generating, um2023don}, self-guided sampling~\citep{um2024self}, or modified noise schedules~\citep{um2025boost}.
Our goal instead is to recover suppressed modes of a \emph{text-conditional} distribution under strong guidance for a fixed prompt.

\subsection{Initial Noise Optimization in Diffusion Models.}
InitNO~\citep{guo2024initno} steers noise toward valid regions via attention-based scores for better text adherence, while \citep{ahn2024noise} learn a mapping to optimal noise configurations to characterize CFG behavior.
Neither addresses diversity.
\del{CNO~\citep{kim2025diverse} is the first to leverage noise optimization for diverse generation, via a pairwise contrastive objective.}\rev{CNO~\citep{kim2025diverse} establishes that optimizing the initial noise alone can recover suppressed modes, using a pairwise contrastive objective.}
We \del{share this initial-noise paradigm but replace the pairwise loss with}\rev{inherit this noise-space pipeline from CNO; what changes is the structure of the diversity signal, from a sum over pairs to} log-determinant maximization, a set-level functional equal to the squared RKHS volume that penalizes redundancy at every order.
This objective is jointly optimized with an anchor-based fidelity term under a single hyperparameter~$\lambda$.
The log-determinant as a diversity criterion originates in GDPP~\citep{elfeki2019gdpp}, which regularizes a GAN's generator with a DPP kernel.
We instead apply it as an inference-time objective on the Tweedie predictions of a frozen diffusion model, requiring no training and no DPP sampling.

\section{Preliminaries}
\label{sec:preliminaries}

\subsection{Tweedie's Formula and Clean Data Estimation}
\label{sec:tweedie_prelim}

Diffusion models generate data by iteratively denoising a Gaussian noise sample $z_T \sim \mathcal{N}(\mathbf{0}, \mathbf{I})$ through a learned reverse process.
At each timestep $t$, a noisy latent $z_t$ is related to the clean data $z_0$ via the forward process $z_t = \sqrt{\bar{\alpha}_t}\, z_0 + \sqrt{1 - \bar{\alpha}_t}\, \epsilon$, where $\epsilon \sim \mathcal{N}(\mathbf{0}, \mathbf{I})$ and $\bar{\alpha}_t \coloneqq \prod_{i=1}^{t} \alpha_i$ is the cumulative noise schedule.
A neural network $\epsilon_\theta(z_t, t, c)$ is trained to predict the noise $\epsilon$ given the noisy observation $z_t$, timestep $t$, and text conditioning $c$.

Tweedie's formula~\citep{Robbins1992} provides a closed-form estimate of the posterior mean $\mathbb{E}[z_0 \mid z_t]$, yielding a one-step prediction of the clean latent at any timestep:
\begin{equation}
    \hat{z}_{0|t}(z_t, t, c) \coloneqq \frac{1}{\sqrt{\bar{\alpha}_t}} \left( z_t - \sqrt{1 - \bar{\alpha}_t} \, \epsilon_\theta(z_t, t, c) \right).
    \label{eq:tweedie}
\end{equation}
This estimate represents the model's best prediction of the clean data given the current noisy observation.
While the prediction becomes increasingly accurate as $t \to 0$, even at the initial timestep $T$ it captures meaningful semantic structure of the final generation, such as global layout and dominant color~\citep{ahn2024noise}.
Our method exploits this property by operating entirely in the Tweedie prediction space at $t = T$, optimizing the initial noises to diversify their corresponding clean data estimates before any denoising takes place.

\subsection{Classifier-Free Guidance}
\label{sec:cfg_prelim}

Classifier-Free Guidance (CFG)~\citep{ho2022classifier} enhances text-image alignment by interpolating between the conditional and unconditional noise predictions:
\begin{equation}
    \hat{\epsilon}_\beta(z_t, t, c) = (1 + \beta)\,\epsilon_\theta(z_t, t, c) 
    - \beta\,\epsilon_\theta(z_t, t, \varnothing),
    \label{eq:cfg}
\end{equation}
where $\beta > 0$ is the guidance scale and $\varnothing$ denotes the null condition. 
This approximately corresponds to sampling from a sharpened distribution $p_\beta(z_0 \mid c) \propto p(z_0)\,p(c \mid z_0)^\beta$, which concentrates probability mass around the dominant modes of the conditional distribution. 
While this yields sharper and more prompt-faithful images, it systematically suppresses low-density modes, reducing output diversity across images generated from the same prompt.

\section{Proposed Method}
\label{sec:method}

\subsection{Constrained Diversity Maximization on Initial Noise}
\label{sec:formulation}

Given a batch of initial latent codes $Z_T = \{z_T^i\}_{i=1}^{B} \sim \mathcal{N}(\mathbf{0}, I)$, we compute one-step Tweedie predictions $\{\hat{z}_{0|T}^i\}_{i=1}^{B}$ via \cref{eq:tweedie}.
Even at the initial timestep $T$, these predictions capture meaningful semantic structure such as global layout, dominant colors, and object placement~\citep{ahn2024noise}.
Since deterministic samplers such as DDIM~\citep{song2020denoising} define a fixed mapping from initial noise to final image, diversifying Tweedie predictions at $t{=}T$ propagates through the entire generation path without per-step intervention.
Our goal is to optimize $Z_T$ so that these predictions are maximally diverse while remaining in plausible regions of the data distribution.
We formalize diversity through RKHS volume maximization and fidelity through an anchor-based displacement constraint, then combine them via Lagrangian relaxation into a single objective.

\begin{algorithm}[t]
\caption{DetNO}
\label{alg:main}
\begin{algorithmic}[1]
\Require Prompt $c$, batch size $B$, fidelity weight $\lambda$, pooling window $w$, steps $N_\mathrm{opt}$, learning rate $\eta$, stability $\epsilon$
\Ensure Optimized initial noise $Z_T^{*}$

\Statex \textit{// Phase 1: Reference computation}
\State Sample $z_T^i \sim \mathcal{N}(\mathbf{0}, I)$ for each $i \in [B]$
\For{$i = 1$ \textbf{to} $B$}
    \State $\hat{z}_{0|T}^{i,\text{ref}} \leftarrow \frac{1}{\sqrt{\bar{\alpha}_T}}\!\left(z_T^i - \sqrt{1-\bar{\alpha}_T}\,\mathrm{stopgrad}\{\epsilon_\theta(z_T^i, T, c)\}\right)$ \Comment{\cref{eq:tweedie}}
    \State $\algpoolref$ \Comment{\cref{eq:pool_norm}}
\EndFor
\State $\algmedian$ \Comment{\cref{eq:median}}

\Statex \textit{// Phase 2: Noise optimization}
\For{$k = 1$ \textbf{to} $N_\mathrm{opt}$}
    \For{$i = 1$ \textbf{to} $B$}
        \State $\hat{z}_{0|T}^{i} \leftarrow \frac{1}{\sqrt{\bar{\alpha}_T}}\!\left(z_T^i - \sqrt{1-\bar{\alpha}_T}\,\mathrm{stopgrad}\{\epsilon_\theta(z_T^i, T, c)\}\right)$ \Comment{\cref{eq:tweedie}}
        \State $\algpoolopt$
    \EndFor
    \State $\alggram$ for all $i, j$ \Comment{\cref{eq:similarity}}
    \State $\algdiv$
    \State $\algfid$
    \State $\algloss$ \Comment{\cref{eq:loss}}
    \State $\algupdate$ for all $i$
\EndFor

\Statex \textit{// Phase 3: Standard sampling}
\State $Z_T^{*} \gets (z_T^1, \ldots, z_T^B)$
\State Generate images from $Z_T^{*}$ using pretrained sampler
\end{algorithmic}
\end{algorithm}

\subsection{Diversity: RKHS Volume Maximization}
\label{sec:diversity}

As motivated in \cref{sec:introduction}, we quantify diversity as the volume a batch collectively occupies, a set-level functional that penalizes redundancy at every order simultaneously.
An unweighted pairwise sum, by contrast, yields a gradient confined to the span of pairwise directions.
Realizing this in the Tweedie prediction space requires care: the raw prediction $\hat{z}_{0|T}^i \in \mathbb{R}^{C \times H \times W}$ suffers from norm concentration, where pairwise distances between independent samples converge to a common value~\citep{aggarwal2001surprising}.
This drives the RBF kernel toward a constant, flattening the very volume signal we maximize.
Predictions at $t{=}T$ primarily encode coarse semantic layout rather than fine spatial detail~\citep{ahn2024noise}.
We therefore suppress high-frequency variance via adaptive average pooling with window $w$, then project onto the unit hypersphere:
\begin{equation}
    \tilde{z}_{0|T}^i = \mathrm{AvgPool}(\hat{z}_{0|T}^i,\; w), \qquad \bar{z}_{0|T}^i = \tilde{z}_{0|T}^i \big/ \|\tilde{z}_{0|T}^i\|_2,
    \label{eq:pool_norm}
\end{equation}
producing compact representations $\bar{z}_{0|T}^i \in \mathbb{R}^{Cw^2}$ on which kernel similarities reflect semantic structure rather than high-frequency noise.
We build the similarity matrix $S \in \mathbb{R}^{B \times B}$ from the RBF kernel, a universal kernel whose Gram matrix is positive semidefinite, so $\log\det(S + \epsilon I)$ is always well-defined:
\begin{equation}
    S_{ij} = \exp\!\left(-\frac{\|\bar{z}_{0|T}^i - \bar{z}_{0|T}^j\|^2}{2\sigma^2}\right), \quad S_{ii} = 1,
    \label{eq:similarity}
\end{equation}
with bandwidth $\sigma$ fixed once from the reference predictions via the median heuristic~\citep{garreau2017large}:
\begin{equation}
    \sigma = \operatorname{median}\!\left(\left\{\|\bar{z}_{0|T}^{i,\mathrm{ref}} - \bar{z}_{0|T}^{j,\mathrm{ref}}\|_2\right\}_{i < j}\right).
    \label{eq:median}
\end{equation}
By the theory of Determinantal Point Processes~\citep{kulesza2012determinantal}, letting $\phi$ be the RBF feature map with $S_{ij} = \langle\phi(\bar{z}_{0|T}^i),\,\phi(\bar{z}_{0|T}^j)\rangle_{\mathcal{H}_k}$,
\begin{equation}
    \det(S) = \mathrm{Vol}^2\!\left(\phi(\bar{z}_{0|T}^1),\,\dots,\,\phi(\bar{z}_{0|T}^B)\right),
    \label{eq:rkhs_volume}
\end{equation}
so $\det(S)$ is the squared volume of the parallelotope spanned by the RKHS embeddings and approaches zero as any embedding approaches the span of the others; for distinct inputs the strictly positive-definite RBF kernel keeps it positive.
Minimizing $-\log\det(S + \epsilon I)$ thus maximizes this volume.
We optimize this volume directly over initial noise rather than sampling from a DPP, which would require intractable normalizing constants over the continuous noise space.
Beyond this geometric interpretation, the log-determinant differs from a sum-of-pairwise objective in how it allocates gradient, concentrating repulsion along the collapsed directions of the batch. We make this precise below.

\begin{proposition}[Spectral Gradient Amplification]
\label{prop:spectral}
Let $\bar{z}^1, \ldots, \bar{z}^B \in \mathbb{S}^{d-1}$ with $d \gg B$, and let $S$ be the $B \times B$ RBF Gram matrix $S_{ij} = \exp(-\|\bar{z}^i - \bar{z}^j\|^2 / 2\sigma^2)$ with eigenvalues $0 < \mu_1 \leq \mu_2 \leq \cdots \leq \mu_B$, writing $\mu_{\min} = \mu_1$.
For $\epsilon > 0$, the Frobenius gradient norms of $\mathcal{L}_{\det} = -\log\det(S + \epsilon I)$ and $\mathcal{L}_{\mathrm{pair}} = \sum_{i < j} S_{ij}$ with respect to $S$ satisfy:
\begin{equation}
    \frac{\|\nabla_S \mathcal{L}_{\det}\|_F}{\|\nabla_S \mathcal{L}_{\mathrm{pair}}\|_F}
    = \frac{2}{\sqrt{B(B-1)}} \sqrt{\sum_{k=1}^{B} \frac{1}{(\mu_k + \epsilon)^2}}.
    \label{eq:spectral_ratio}
\end{equation}
In particular: \emph{(i)} the ratio is bounded below by $2 / \bigl(\sqrt{B(B-1)}\,(\mu_1 + \epsilon)\bigr)$; \emph{(ii)} the ratio exceeds $1$ whenever $\mu_1 < 2/\sqrt{B(B-1)} - \epsilon$ (sufficient condition via bound~\emph{(i)}); \emph{(iii)} the ratio is self-attenuating: as all eigenvalues grow during optimization, the sum in \cref{eq:spectral_ratio} decreases monotonically, concentrating gradient energy where redundancy is highest and relaxing repulsion as diversity is achieved.
\end{proposition}
\noindent
The proof is in \cref{sec:proof_spectral}; the identity in \cref{eq:spectral_ratio} holds for any positive-definite Gram matrix, with $d \gg B$ serving only as a non-degeneracy condition.
The proposition characterizes the ratio in Gram-matrix space; empirically the amplification carries over to the noise-space gradient (Pearson $r{=}0.999$, \del{Table~5}\rev{\cref{sec:proof_spectral}}), so the repulsion intensifies as the batch collapses and relaxes once volume is restored.

\subsection{Fidelity: Anchor-Based Displacement Constraint}
\label{sec:fidelity}

Volume maximization alone would push noise vectors into out-of-distribution regions.
Since CFG concentrates $p_\beta(z_0 \mid c) \propto p(z_0)\,p(c \mid z_0)^\beta$ around high-density modes, each reference prediction $\hat{z}_{0|T}^{i,\mathrm{ref}}$ from the unoptimized noise lies near a high-density region of the sharpened posterior by construction.
We regularize displacement from these anchors in the full-resolution prediction space:
\begin{equation}
    \left\|\hat{z}_{0|T}^i - \hat{z}_{0|T}^{i,\mathrm{ref}}\right\|^2 \leq \delta, \quad \forall\, i \in \{1,\dots,B\}.
    \label{eq:constraint}
\end{equation}
The constrained problem is therefore:
\begin{equation}
    \min_{Z_T}\; -\log\det(S + \epsilon I) \quad \mathrm{s.t.} \quad \left\|\hat{z}_{0|T}^i - \hat{z}_{0|T}^{i,\mathrm{ref}}\right\|^2 \leq \delta, \quad \forall\, i \in \{1,\dots,B\}.
    \label{eq:constrained}
\end{equation}
Introducing per-sample Lagrange multipliers $\{\lambda_i\}_{i=1}^B$ ($\lambda_i \geq 0$) gives the Lagrangian:
\begin{equation}
    \mathcal{L} = -\log\det(S + \epsilon I) + \sum_{i=1}^{B} \lambda_i \left(\left\|\hat{z}_{0|T}^i - \hat{z}_{0|T}^{i,\mathrm{ref}}\right\|^2 - \delta\right).
    \label{eq:lagrangian_general}
\end{equation}
As a practical simplification, since all $B$ samples are drawn from the same prompt, backbone, and noise prior, we use a single shared multiplier $\lambda$ across all samples and absorb the constant $-\lambda B\delta$ into $\lambda$, arriving at the final objective:
\begin{equation}
    \boxed{
    \mathcal{L} = \underbrace{-\log\det(S + \epsilon I)}_{\text{Diversity}} + \underbrace{\lambda \sum_{i=1}^{B} \left\| \hat{z}_{0|T}^i - \hat{z}_{0|T}^{i,\mathrm{ref}} \right\|^2}_{\text{Fidelity}}
    }
    \label{eq:loss}
\end{equation}
where $\epsilon > 0$ ensures numerical stability.
The two terms are fully decoupled: diversity depends on pooled, normalized representations, while fidelity operates on unpooled predictions at full spatial resolution.
A single hyperparameter $\lambda$ transparently controls the trade-off, with no sampler modification required.
Following~\citep{kim2025diverse}, we apply a stop-gradient operator on the UNet output when computing Tweedie predictions.
This eliminates the need to store UNet activations during backpropagation, reducing GPU memory to the level of standard inference.
\cref{alg:main} summarizes the full procedure.

\section{Experiments}
\label{sec:experiments}

\begin{figure}[t!]
    \centering
    \includegraphics[width=1\linewidth]{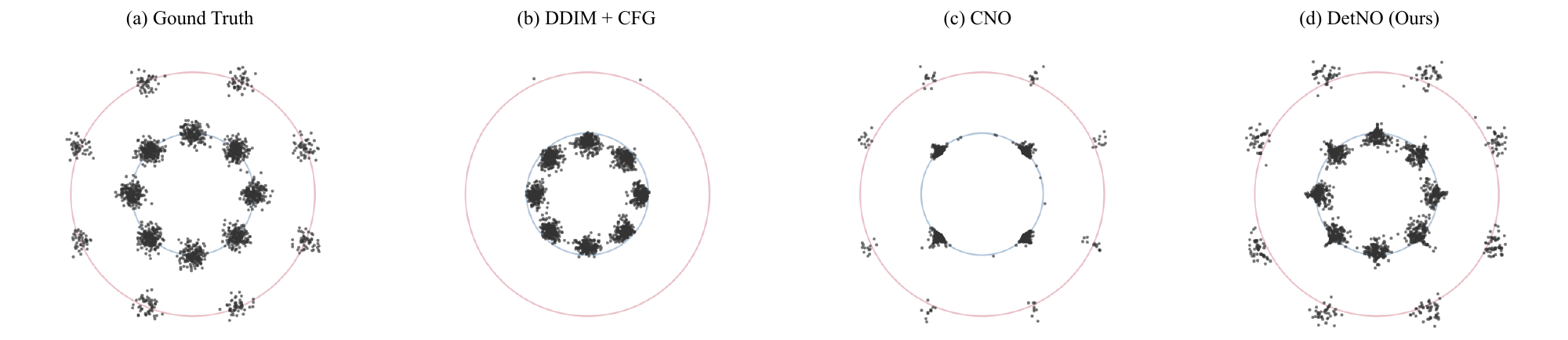}
    \caption{\textbf{Toy example on a 2D ring mixture} with 8 inner modes (80\%) and 8 outer modes (20\%). (a) Ground truth distribution. (b) Classifier-free guidance collapses onto the majority inner ring, without covering outer modes. (c) CNO recovers the outer modes but severely distorts the inner ring, producing only 4 inner modes whose sample distributions no longer align with the ground truth geometry. (d) DetNO achieves full coverage of all 16 modes across both rings, faithfully preserving the original mode structure.}
    \label{fig:toy_example}
\end{figure}

\paragraph{Implementation Details.}
We evaluate on three pre-trained T2I backbones: SD1.5~\citep{rombach2022high}, SDXL~\citep{podell2023sdxl}, and SD3~\citep{esser2024scaling}.
SD1.5 and SDXL use DDIM~\citep{song2020denoising} with 50 steps, while SD3 uses the Euler sampler with 28 steps following its default configuration.
We compare against PG~\citep{corso2023particle}, SPELL~\citep{kirchhof2024shielded}, DiversityPrompt~\citep{um2025minority}, and CNO~\citep{kim2025diverse}.
We evaluate all methods under identical conditions (prompt set, random seeds, CFG scale, and hardware) to ensure fair relative comparison; absolute values may differ from original papers accordingly.
We use the official implementations of all baselines, including CNO, with the hyperparameters reported in their original papers.
Following~\citep{um2025minority, kim2025diverse}, we sample 2,000 prompts from the MS-COCO~\citep{lin2014microsoft} validation set and generate 5 images per prompt for SD1.5 and SD3, and 3 images per prompt for SDXL.

\paragraph{Evaluation Metrics.}
For image quality, we report Precision~\citep{kynkaanniemi2019improved},
Density~\citep{naeem2020reliable}, CLIPScore~\citep{hessel2021clipscore},
PickScore~\citep{kirstain2023pick}, and ImageReward~\citep{xu2023imagereward}.
For diversity, we report Recall~\citep{kynkaanniemi2019improved},
Coverage~\citep{naeem2020reliable}, Vendi Score~\citep{friedman2022vendi},
and Mean Self-Similarity (MSS)~\citep{kim2025diverse}.
Among these, VendiScore and MSS serve as our primary diversity metrics, as they measure intra-batch coverage directly without reference to an external distribution.
They are therefore better suited to evaluating diversity-enhancing methods than reference-dependent metrics such as Recall and Coverage~\citep{park2023probabilistic}.

\subsection{Results}
\begin{table*}[t!]
\centering
\caption{Quantitative comparison across different backbone models.
\textbf{Bold} and \underline{underline} denote the best and second-best results, respectively.}
\label{tab:main_results}
\resizebox{\textwidth}{!}{%
\begin{tabular}{ll | cccc | ccc | cc}
\toprule
& & \multicolumn{4}{c|}{\textbf{Reference-Based}} & \multicolumn{3}{c|}{\textbf{Reference-Free}} & \multicolumn{2}{c}{\textbf{Diversity}} \\
\textbf{Model} & \textbf{Method} & \textbf{Prec $\uparrow$} & \textbf{Rec $\uparrow$} & \textbf{Den $\uparrow$} & \textbf{Cov $\uparrow$} & \textbf{CLIP $\uparrow$} & \textbf{Pick $\uparrow$} & \textbf{IR $\uparrow$} & \textbf{MSS $\downarrow$} & \textbf{Vendi $\uparrow$} \\
\midrule
\multirow{6}{*}{SD1.5}
 & DDIM & 0.7005 & 0.6061 & 0.6112 & 0.3631 & 30.9096 & \underline{21.3413} & 0.1086 & 0.1509 & 4.7352 \\
 & PG & \underline{0.7008} & 0.6044 & \underline{0.6120} & \underline{0.3634} & 30.9103 & \textbf{21.3424} & 0.1085 & 0.1508 & 4.7357 \\
 & SPELL & 0.6722 & 0.5645 & 0.5639 & 0.3439 & 30.7538 & 21.2556 & 0.0264 & 0.1409 & 4.7636 \\
 & DiversityPrompt & 0.6679 & \underline{0.6442} & 0.5345 & 0.3370 & \underline{30.9380} & 21.1956 & \textbf{0.1194} & 0.1455 & 4.7509 \\
 & CNO & \textbf{0.7247} & 0.6318 & \textbf{0.6647} & \textbf{0.3933} & 30.7925 & 21.2984 & 0.0702 & \underline{0.1383} & \underline{4.7679} \\
\rowcolor{oursblue} & Ours & \rev{0.6744} & \rev{\textbf{0.6807}} & \rev{0.5541} & \rev{0.3570} & \rev{\textbf{31.1396}} & \rev{21.2536} & \rev{\underline{0.1088}} & \rev{\textbf{0.1229}} & \rev{\textbf{4.8024}} \\
\midrule
\multirow{5}{*}{SDXL}
 & DDIM & 0.6623 & 0.6677 & \underline{0.5399} & \underline{0.2466} & \textbf{31.8141} & \underline{22.5280} & \underline{0.7784} & 0.2031 & 2.8560 \\
 & PG & \underline{0.6708} & 0.6821 & 0.5354 & \underline{0.2466} & \underline{31.7768} & \textbf{22.5375} & \textbf{0.7864} & 0.2007 & 2.8589 \\
 & SPELL & \textbf{0.6852} & 0.6690 & \textbf{0.5545} & \textbf{0.2525} & 31.7059 & 22.4448 & 0.7296 & 0.1902 & 2.8712 \\
 & CNO & 0.6592 & \underline{0.6824} & 0.5169 & 0.2446 & 31.7637 & 22.4910 & 0.7644 & \underline{0.1869} & \underline{2.8746} \\
\rowcolor{oursblue} & Ours & 0.6552 & \textbf{0.6945} & 0.4966 & 0.2362 & 31.7273 & 22.4489 & 0.7443 & \textbf{0.1720} & \textbf{2.8920} \\
\midrule
\multirow{5}{*}{SD3}
 & Euler & 0.6261 & 0.5223 & 0.4527 & 0.2480 & \textbf{31.8805} & 22.5719 & \underline{1.0503} & 0.2944 & 4.2517 \\
 & PG & 0.6334 & 0.5238 & 0.4456 & 0.2487 & \underline{31.8631} & \underline{22.5806} & \textbf{1.0510} & 0.2947 & 4.2485 \\
 & SPELL & \textbf{0.6799} & 0.5172 & \textbf{0.5241} & \textbf{0.2761} & 31.8447 & \textbf{22.5992} & 1.0362 & 0.2866 & 4.2791 \\
 & CNO & 0.6396 & \underline{0.5372} & 0.4495 & 0.2518 & 31.8481 & 22.5512 & 1.0423 & \underline{0.2760} & \underline{4.3196} \\
\rowcolor{oursblue} & Ours & \underline{0.6427} & \textbf{0.5441} & \underline{0.4601} & \underline{0.2648} & 31.8567 & 22.4967 & 1.0109 & \textbf{0.2542} & \textbf{4.3988} \\
\bottomrule
\end{tabular}
}
\end{table*}

\paragraph{Toy Example: 2D Ring Mixture.}
Only the log-determinant preserves the full mode structure in a controlled setting, a 2D ring mixture with 8 inner modes (80\%) and 8 outer modes (20\%) (\cref{fig:toy_example}).
CFG (b) collapses onto the majority inner ring, CNO (c) recovers the outer ring while distorting the inner one, and DetNO (d) covers all 16 modes.
The two failures are distinct: CFG cannot reach the suppressed modes at all, while InfoNCE-style contrastive repulsion reaches them but keeps pushing after samples are already separated, distorting the geometry it recovers.
The log-determinant instead self-attenuates as volume is restored (\cref{prop:spectral}), separating samples without displacing the modes themselves.

\begin{figure}[t!]
    \centering
    \includegraphics[width=1\linewidth]{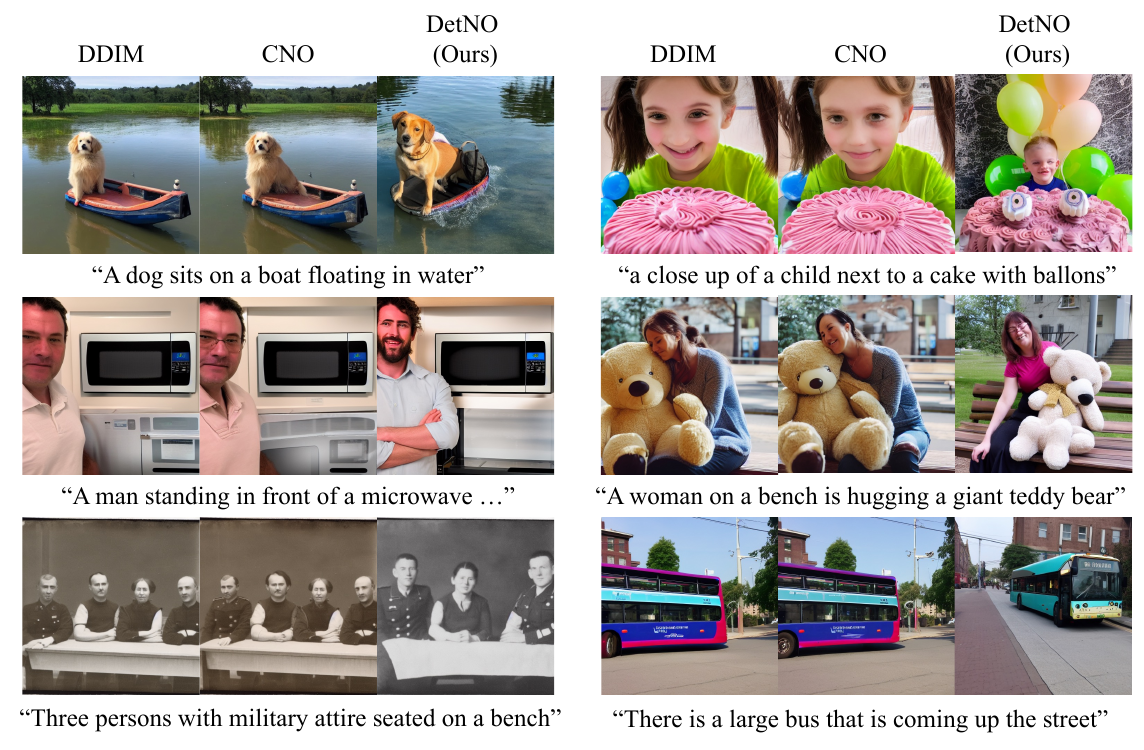}
    \caption{\textbf{Qualitative comparison on SD1.5.}
    For each prompt, we show one representative sample from DDIM, CNO, and DetNO (ours).
    DDIM and CNO produce visually near-identical outputs across seeds, differing only in minor details such as slight pose or hair difference.
    DetNO generates substantially different images\del{ —}\rev{,} varying in viewpoint, object appearance, scene layout, and human identity\del{ —}\rev{,} while remaining faithful to the prompt.}
    \label{fig:qualitative_sd15}
\end{figure}

\begin{table}[t!]
\centering
\caption{\textbf{Downstream few-shot classification accuracy (\%).}
Mean $\pm$ std over 10 splits.
Synthetic images (16/class) generated with SD3 (Oxford-IIIT Pets) and SD1.5 (CUB-200-2011).
$\dagger$ and $\ddagger$ denote statistically significant improvement over DDIM and CNO respectively (Welch's $t$-test, $p < 0.01$).}
\label{tab:downstream}
\resizebox{\linewidth}{!}{
\begin{tabular}{lcccccccc}
\toprule
& \multicolumn{4}{c}{Oxford-IIIT Pets (SD3)} & \multicolumn{4}{c}{CUB-200-2011 (SD1.5)} \\
\cmidrule(lr){2-5} \cmidrule(lr){6-9}
Method & 1-shot & 2-shot & 4-shot & 8-shot & 1-shot & 2-shot & 4-shot & 8-shot \\
\midrule
Real only
  & 17.6 $\pm$ 2.0 & 41.2 $\pm$ 2.9 & 57.7 $\pm$ 2.7 & 65.4 $\pm$ 0.7
  & 8.2 $\pm$ 0.5 & 15.1 $\pm$ 0.6 & 27.9 $\pm$ 0.9 & 47.0 $\pm$ 1.3 \\
+ DDIM
  & 52.2 $\pm$ 1.8 & 58.2 $\pm$ 0.9 & 59.1 $\pm$ 1.3 & \textbf{71.8 $\pm$ 0.7}
  & 24.6 $\pm$ 0.9 & 31.1 $\pm$ 0.9 & 38.5 $\pm$ 0.9 & 50.9 $\pm$ 0.8 \\
+ CNO
  & 51.7 $\pm$ 1.7 & 57.8 $\pm$ 0.6 & 59.4 $\pm$ 1.6 & 71.4 $\pm$ 0.7
  & 24.5 $\pm$ 0.7 & 30.6 $\pm$ 0.8 & 38.4 $\pm$ 0.5 & 50.7 $\pm$ 1.1 \\
+ DetNO
  & \textbf{54.2 $\pm$ 1.1}$^{\ddagger}$ & \textbf{59.3 $\pm$ 1.3}$^{\ddagger}$ & \textbf{61.2 $\pm$ 1.7}$^{\dagger}$ & \textbf{71.8 $\pm$ 0.9}
  & \textbf{26.3 $\pm$ 0.5}$^{\dagger\ddagger}$ & \textbf{32.1 $\pm$ 0.7}$^{\ddagger}$ & \textbf{39.8 $\pm$ 0.6}$^{\dagger\ddagger}$ & \textbf{51.8 $\pm$ 0.9} \\
\bottomrule
\end{tabular}}
\end{table}

\begin{figure}[t!]
    \centering
    \includegraphics[width=0.8\linewidth]{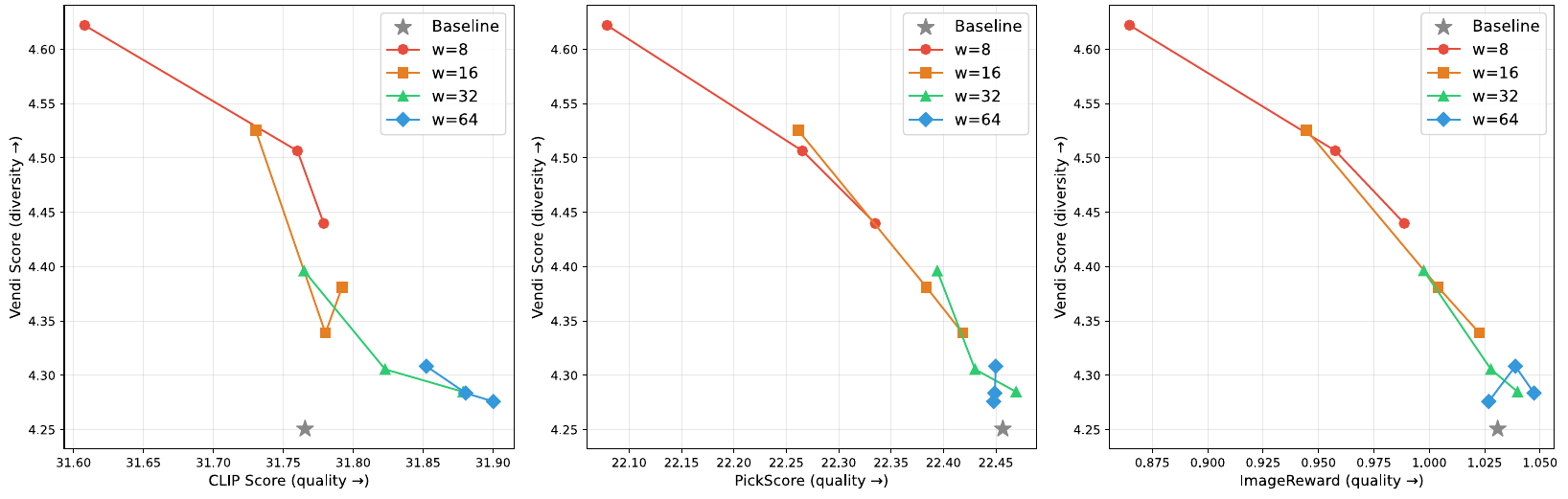}
    \caption{\textbf{Quality--diversity Pareto curves on SD3} for different pooling windows $w \in \{8, 16, 32, 64\}$.
    Each point on a curve corresponds to a different fidelity weight $\lambda \in \{0.01, 0.05, 0.1\}$, evaluated on 200 randomly sampled MS-COCO prompts with 5 images per prompt.
    All configurations \del{consistently dominate}\rev{improve over} the unoptimized baseline (gray star) \rev{on this evaluation}.
    Smaller $w$ yields a wider diversity range at the cost of a steeper quality trade-off, while larger $w$ preserves quality more effectively with moderate diversity gains.}
    \label{fig:w_lambda_pareto}
\end{figure}

\paragraph{Quantitative Results.}
DetNO achieves the best MSS and VendiScore on all three (SD1.5, SDXL, SD3) in \cref{tab:main_results}, confirming systematic diversity gains across all three backbones and both samplers tested (DDIM and Euler).
\del{CNO's improvements, by contrast, are competitive on SD1.5 but marginal on SDXL and SD3.}\rev{CNO also improves diversity, but its VendiScore gain over the baseline is roughly half of DetNO's on each backbone.}
On reference-free quality metrics (PickScore, ImageReward), the trade-off is \del{backbone-dependent: negligible on SD3 but more pronounced on SD1.5 and SDXL}\rev{small: PickScore decreases only marginally on every backbone, while ImageReward matches the baseline on SD1.5 and decreases modestly on SDXL and SD3}.
\del{Recovering modes outside the high-density region inherently shifts the operating point along the quality--diversity frontier.}
Reference-dependent metrics (Precision, Density, Coverage) decrease accordingly, as they reward overlap with the narrow MS-COCO reference set~\citep{park2023probabilistic} rather than per-image quality.
The trade-off remains controllable: the Pareto analysis (\cref{fig:w_lambda_pareto}) shows $\lambda$ smoothly interpolating between the two objectives, and on \del{SD3 every tested configuration dominates the baseline across CLIPScore, PickScore, and ImageReward}\rev{its 200-prompt sweep every tested SD3 configuration improves over the baseline on CLIPScore, PickScore, and ImageReward; under the full 2,000-prompt protocol, the deployed configuration trades a small amount of PickScore and ImageReward for its diversity gains (\cref{tab:main_results})}.

\paragraph{Qualitative Results.}
The diversity gains are plainly visible (\cref{fig:qualitative_sd15}, six SD1.5 prompts; SDXL and SD3 examples in \cref{fig:supple_sdxl,fig:supple_sd3}).
DDIM and CNO are nearly indistinguishable, sharing global layout, object placement, and dominant color palette across all examples.
DetNO instead varies object identity, viewpoint, and scene composition while remaining prompt-faithful, recovering degrees of freedom in the initial noise that standard sampling and pairwise objectives leave unexploited.

\paragraph{Downstream Classification.}
These diversity gains translate into practical utility.
We evaluate synthetic data augmentation in a few-shot classification setting using two fine-grained benchmarks: Oxford-IIIT Pets (37 breeds)~\citep{parkhi12a} with SD3 and CUB-200-2011 (200 species)~\citep{WahCUB_200_2011} with SD1.5.
For each class, we generate 16 synthetic images and combine them with $K$-shot real images to fine-tune a pretrained ResNet-18 (Tab.~\ref{tab:downstream}).
DetNO consistently outperforms both DDIM and CNO across both datasets, with gains most pronounced in the low-data regime: $+2.0$pp at 1-shot on Pets and $+1.7$pp on CUB-200 over DDIM, where distributional breadth of synthetic data is directly impactful.
\del{CNO does not improve over DDIM on either dataset}\rev{CNO remains within noise of DDIM on both datasets}, whereas DetNO's broader mode coverage provides tangible downstream benefits beyond what diversity metrics alone would suggest.

\begin{table}[t!]
\centering
\caption{\textbf{Pairwise-RBF ablation with downstream evaluation.}
SD1.5, $B\!=\!5$, $\lambda\!=\!0.05$; $S$, $\sigma$, fidelity term, and optimizer are held identical across rows, so only the batch aggregator differs: InfoNCE (CNO), the sum-of-pairwise RBF $\sum_{i<j} S_{ij}$, and the log-determinant (DetNO). Downstream columns follow the protocol of \cref{tab:downstream}. Bold and underline denote the best and second-best results.}
\label{tab:pairwise_rbf}
\resizebox{0.8\linewidth}{!}{%
\begin{tabular}{lcccc}
\toprule
\multirow{2}{*}{Method} & \multicolumn{2}{c}{SD1.5 (intra-batch)} & \multicolumn{2}{c}{1-shot accuracy (\%)} \\
\cmidrule(lr){2-3}\cmidrule(lr){4-5}
 & Pick $\uparrow$ & \del{VendiScore}\rev{Vendi} $\uparrow$ & Pets (SD3) $\uparrow$ & CUB (SD1.5) $\uparrow$ \\
\midrule
DDIM (Baseline) & \rev{\textbf{21.34}} & \rev{4.7352} & 52.2 \scriptsize{$\pm$ 1.8} & 24.6 \scriptsize{$\pm$ 0.9} \\
CNO (InfoNCE) & \underline{21.30} & 4.7679 & 51.7 \scriptsize{$\pm$ 1.7} & 24.5 \scriptsize{$\pm$ 0.7} \\
$\sum_{i<j} S_{ij}$ (pairwise-RBF) & 21.23 & \underline{4.8006} & \underline{53.7 \scriptsize{$\pm$ 1.1}} & \underline{25.8 \scriptsize{$\pm$ 0.6}} \\
$-\log\det$ (DetNO) & 21.25 & \textbf{4.8024} & \textbf{54.2 \scriptsize{$\pm$ 1.1}} & \textbf{26.3 \scriptsize{$\pm$ 0.5}} \\
\bottomrule
\end{tabular}%
}
\end{table}

\paragraph{Isolating the Determinant Structure.}
Holding $S$, $\sigma$, the fidelity term, and the optimizer fixed, we vary only the batch aggregator among InfoNCE (CNO), the sum-of-pairwise RBF $\sum_{i<j} S_{ij}$, and our log-determinant (\cref{tab:pairwise_rbf}), isolating the determinant structure from the inherited pipeline.
The log-determinant attains the best intra-batch diversity, and both RBF-based aggregators outperform InfoNCE.
Because the pairwise-RBF and log-determinant rows differ only in aggregating $S$ additively versus through its determinant, the comparison isolates the aggregation structure.
The gap at $B{=}5$ is small, as expected where higher-order redundancies are few; most of the gain over InfoNCE comes from the shared RBF kernel and fidelity design.
We report intra-batch diversity for comparability but treat downstream accuracy as decisive.
There the advantage is sharpest: the log-determinant reaches the highest 1-shot accuracy on both benchmarks, while InfoNCE alone does not surpass the DDIM baseline.

\subsection{Ablation Studies}

\paragraph{Effect of pooling window $w$ and fidelity weight $\lambda$.}
The method is robust to the choice of $w$ on SD3: \cref{fig:w_lambda_pareto} traces quality--diversity frontiers on SD3 by sweeping $\lambda \in \{0.01, 0.05, 0.1\}$ for each pooling window $w \in \{8, 16, 32, 64\}$.
\del{All configurations consistently dominate the unoptimized baseline across CLIPScore, PickScore, and ImageReward.}\rev{On this 200-prompt evaluation, all configurations improve over the unoptimized baseline across CLIPScore, PickScore, and ImageReward.}
Smaller $w$ operates on coarser semantic features, giving a wider diversity range but a steeper quality slope; larger $w$ preserves finer spatial structure and trades less quality per unit of diversity gain.
We select $w{=}32$ for our SD3 experiments as it provides a favorable balance across all three quality metrics.

\begin{table}[t]
  \centering
  \caption{Ablation study on the loss components of \cref{eq:loss} in the SD1.5. A: diversity term, B: fidelity term.}
  \label{tab:ablation}
  \resizebox{0.8\linewidth}{!}{
  \begin{tabular}{ccccccccccc}
    \toprule
    A & B & Prec$\uparrow$ & Recall$\uparrow$ & Den$\uparrow$ & Cov$\uparrow$ & CLIP$\uparrow$ & Pick$\uparrow$ & IR$\uparrow$ & MSS$\downarrow$ & \del{VS}\rev{Vendi}$\uparrow$ \\
    \midrule
    && \underline{0.7005} & 0.6061 & \textbf{0.6112} & \textbf{0.3631} & 30.9096 & \underline{21.3413} & 0.1086 & 0.1509 & 4.7352 \\
    \cmark & & 0.6816 & \underline{0.6191} & 0.5686 & 0.3468 & 30.7146 & 21.1857 & -0.0489 & \textbf{0.1201} & \textbf{4.8107} \\
     & \cmark & \textbf{0.7014} & 0.6024 & \underline{0.5924} & 0.3531 & \underline{30.9112} & \textbf{21.3459} & \textbf{0.1187} & 0.1516 & 4.7334 \\
    \rowcolor{oursblue} \cmark & \cmark & \rev{0.6744} & \rev{\textbf{0.6807}} & \rev{0.5541} & \rev{\underline{0.3570}} & \rev{\textbf{31.1396}} & \rev{21.2536} & \rev{\underline{0.1088}} & \rev{\underline{0.1229}} & \rev{\underline{4.8024}} \\
    \bottomrule
  \end{tabular}
  }
\end{table}

\begin{figure}[t!]
    \centering
    \includegraphics[width=1\linewidth]{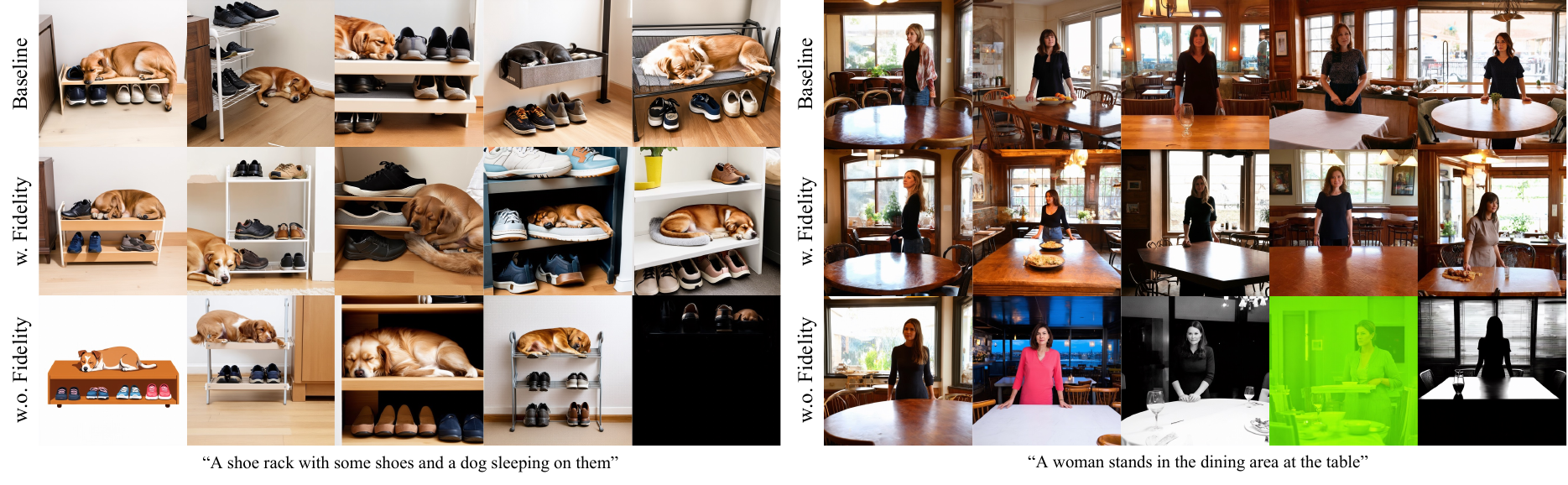}
    \caption{\textbf{Qualitative ablation of the fidelity term on SD3.} Top: unoptimized baseline. With fidelity (middle), DetNO increases diversity while preserving image quality. Without fidelity (bottom), unconstrained volume maximization pushes noise vectors into out-of-distribution regions, producing artifacts such as cartoon-like renderings, color corruption, and degenerate outputs.}
    \label{fig:fidelity_ablation}
\end{figure}

\begin{figure}[t!]
    \centering
    \includegraphics[width=0.8\linewidth]{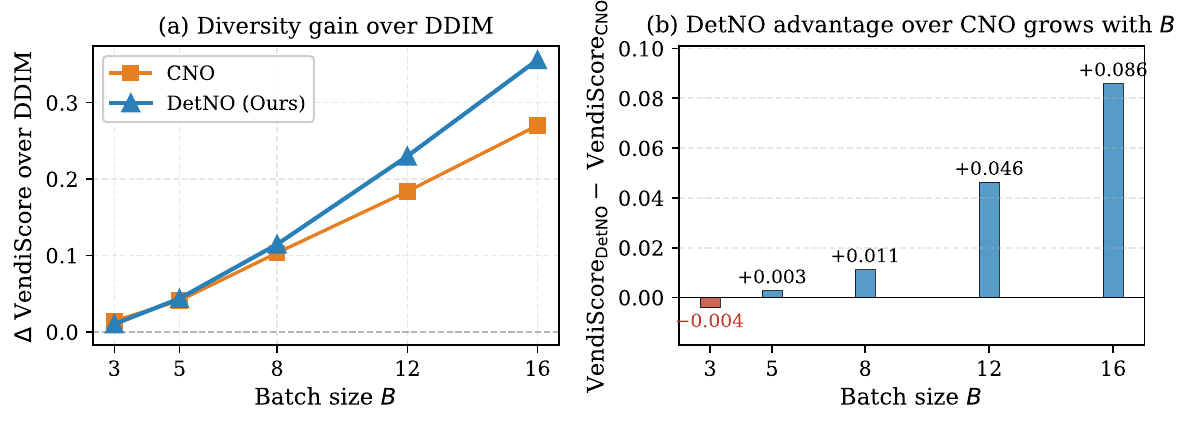}
    \caption{\textbf{Batch-size scaling on SD1.5.}
    (a) VendiScore gain over the DDIM baseline versus batch size $B$ for CNO and DetNO.
    (b) DetNO's VendiScore advantage over CNO, which widens monotonically with $B$\del{, consistent with \cref{prop:spectral}}.
    All other settings follow \cref{tab:hyperparams}.}
    \label{fig:batch_ablation}
\end{figure}

\paragraph{Ablation study on the loss components.}
Both loss terms are necessary: \cref{tab:ablation} isolates each term's contribution on SD1.5.
The diversity term alone (A) achieves the strongest diversity but at \del{significant}\rev{substantial} quality cost, pushing noise vectors into out-of-distribution regions (bottom row of \cref{fig:fidelity_ablation}).
The fidelity term alone (B) preserves quality but provides no diversity improvement.
The full objective (A+B) recovers most diversity gains while maintaining prompt-faithful outputs (middle row of \cref{fig:fidelity_ablation}).

\paragraph{Scaling with batch size.}
\del{A single batch size understates the gap between all-order and pairwise objectives, which}\rev{All-order and pairwise objectives} coincide at $B\!=\!2$ and diverge only as higher-order redundancies accumulate; we therefore sweep $B \in \{3,5,8,12,16\}$ on SD1.5 with all else fixed (\cref{fig:batch_ablation}).
Both methods improve over DDIM, but DetNO's advantage over CNO is marginal at small $B$ and widens monotonically as $B$ grows.
The trend \del{matches}\rev{is consistent with} the all-order argument: a pairwise objective sees only the $\binom{B}{2}$ pair terms, while the higher-order redundancies the log-determinant penalizes grow combinatorially with $B$.
\rev{As the two methods also differ in kernel and fidelity design, the gap reflects their combined effect.}
Its gradient adds an $O(B^3)$ matrix factorization per step, negligible against the $B$ denoiser evaluations that dominate each iteration.
Synthetic data augmentation operates in exactly this large-batch regime.

\section{Conclusion}
We presented DetNO, an initial-noise optimization framework that maximizes the RKHS volume among Tweedie predictions via a log-determinant diversity term and an anchor-based fidelity term.
\del{A single hyperparameter $\lambda$ provides transparent control over the quality--diversity trade-off, with no sampler modification.}
Experiments on SD1.5, SDXL, and SD3 confirm consistent diversity gains across backbones and samplers, with Pareto analysis showing that $\lambda$ enables smooth interpolation between the two objectives.
A toy experiment and a spectral analysis (\cref{prop:spectral}) trace this advantage to its mechanism: repulsion that intensifies where the batch is most collapsed and relaxes once diversity is restored.
In downstream few-shot classification, the broader mode coverage translates into statistically significant accuracy gains in the low-data regime.
Treating diversity as a property of the whole batch, rather than of its pairs, thus produces image sets that are not only more varied but measurably more useful.

\section*{Acknowledgments}
This was partly supported by the Institute of Information \& Communications Technology Planning \& Evaluation (IITP) grant funded by the \grantsponsor{Korean government (MSIT)} (No. \grantnumber{1}{RS-2020-II201373}, Artificial Intelligence Graduate School Program(Hanyang University)) and the Institute of Information \& Communications Technology Planning \& Evaluation (IITP) grant funded by the \grantsponsor{Korea government (MSIT)} (No. \grantnumber{2}{RS-2025-02219062}, Self-training framework for VLM-based defect detection and explanation model in manufacturing process).

\bibliographystyle{splncs04}
\bibliography{main}

\clearpage
\appendix
\setcounter{figure}{0}
\setcounter{table}{0}
\setcounter{equation}{0}
\renewcommand{\thetable}{S\arabic{table}}
\renewcommand{\thefigure}{S\arabic{figure}}
\renewcommand{\theequation}{S\arabic{equation}}
\section{Appendix}
\subsection{Proof of \cref{prop:spectral}}
\label{sec:proof_spectral}

We first note that the algebraic identity in \cref{eq:spectral_ratio} holds for \emph{any}
$B \times B$ positive-definite RBF Gram matrix, regardless of the ambient dimension $d$.
Two separate conditions on $S$ are relevant here.
First, the RBF kernel $k(x,y) = \exp(-\|x-y\|^2/2\sigma^2)$ is \emph{strictly} positive
definite for any finite set of distinct points in any dimension,
ensuring that $S$ is always invertible and that $\log\det(S+\epsilon I)$ is well-defined
without any assumption on $d$.
Second, the assumption $d \gg B$ serves as a \emph{non-degeneracy} condition for the
eigenvalue bounds: when $d \gg B$, generic points on $\mathcal{S}^{d-1}$ are in general
position with probability one, ensuring that the eigenvalues $\{\mu_k\}$ are bounded
away from zero and that the bounds in \cref{prop:spectral} are non-vacuous.

Using the matrix identity $\nabla_S \log\det(M) = M^{-1}$ for any symmetric
positive-definite $M$, the gradient of $\mathcal{L}_{\det}$ with respect to $S$ is:
\begin{equation}
    \nabla_S \mathcal{L}_{\det} = -(S + \epsilon I)^{-1}.
\end{equation}
Since $S$ admits the eigendecomposition $S = U \operatorname{diag}(\mu_1, \ldots, \mu_B)
U^\top$, the inverse $(S + \epsilon I)^{-1}$ has eigenvalues $\{1/(\mu_k +
\epsilon)\}_{k=1}^B$. Since the Frobenius norm of a symmetric matrix equals the $\ell_2$
norm of its eigenvalues:
\begin{equation}
    \|\nabla_S \mathcal{L}_{\det}\|_F
    = \|(S + \epsilon I)^{-1}\|_F
    = \sqrt{\sum_{k=1}^{B} \frac{1}{(\mu_k + \epsilon)^2}}.
    \label{eq:grad_det_norm}
\end{equation}
For the pairwise objective, rewriting $\mathcal{L}_{\mathrm{pair}} =
\frac{1}{2}(\mathbf{1}^\top S \mathbf{1} - \operatorname{tr}(S))$ and differentiating
with respect to the symmetric matrix $S$ yields:
\begin{equation}
    \nabla_S \mathcal{L}_{\mathrm{pair}} = \tfrac{1}{2}(J - I), \quad J =
    \mathbf{1}\mathbf{1}^\top.
\end{equation}
This matrix has zero diagonal and $B(B-1)$ off-diagonal entries each equal to
$\frac{1}{2}$, giving:
\begin{equation}
    \|\nabla_S \mathcal{L}_{\mathrm{pair}}\|_F
    = \sqrt{B(B-1) \cdot \left(\tfrac{1}{2}\right)^2}
    = \tfrac{1}{2}\sqrt{B(B-1)}.
    \label{eq:grad_pair_norm}
\end{equation}
Dividing \cref{eq:grad_det_norm} by \cref{eq:grad_pair_norm} then directly yields:
\begin{equation}
    \frac{\|\nabla_S \mathcal{L}_{\det}\|_F}{\|\nabla_S \mathcal{L}_{\mathrm{pair}}\|_F}
    = \frac{\sqrt{\displaystyle\sum_{k=1}^{B} \frac{1}{(\mu_k + \epsilon)^2}}}
           {\dfrac{1}{2}\sqrt{B(B-1)}}
    = \frac{2}{\sqrt{B(B-1)}} \sqrt{\sum_{k=1}^{B} \frac{1}{(\mu_k + \epsilon)^2}},
\end{equation}
which is exactly \cref{eq:spectral_ratio}. Bound~(i) follows from dropping all but the
smallest-eigenvalue term: $\sum_k 1/(\mu_k+\epsilon)^2 \geq 1/(\mu_1+\epsilon)^2$. For
bound~(ii), applying bound~(i) and imposing the ratio to exceed unity gives:
\begin{equation}
    \frac{2}{\sqrt{B(B-1)}\,(\mu_1 + \epsilon)} > 1
    \quad\Longleftrightarrow\quad
    \mu_1 < \frac{2}{\sqrt{B(B-1)}} - \epsilon,
\end{equation}
and substituting $B=5$, $\epsilon=10^{-5}$ gives the threshold $2/\sqrt{20} - 10^{-5}
\approx 0.4472$. Bound~(iii) follows since $\sum_k 1/(\mu_k+\epsilon)^2$ is strictly
decreasing in each $\mu_k$, so the ratio self-attenuates monotonically as eigenvalues
grow during optimization, concentrating gradient energy where redundancy is highest and
relaxing repulsion as diversity is achieved.
\qed

\begin{remark}[$S$-level vs.\ $z$-level gradient amplification]
\label{rem:z_level}
\cref{prop:spectral} bounds the ratio of gradient norms with respect to the Gram matrix
$S$, whereas the empirically measured ratio $\|\nabla_z \mathcal{L}_{\det}\|_F /
\|\nabla_z \mathcal{L}_{\mathrm{pair}}\|_F$ additionally depends on the Jacobian
$\partial S / \partial z$ via the chain rule. The strong empirical coupling (Pearson $r =
0.999$ at mean level, $R^2 = 0.999$ for the linear fit $R_z =
0.173/(\mu_{\min}+\epsilon) + 0.394$; \cref{tab:grad_dynamics}) confirms that the
$S$-level amplification mechanism propagates faithfully through this Jacobian to the
actual optimization variables.
\end{remark}

\subsection{Derivation of the Two-Term Objective via Lagrangian Relaxation}
\label{sec:lagrangian_derivation}

Our primary objective is to maximize batch diversity in the RKHS of the Tweedie prediction space by minimizing $-\log\det(S+\epsilon I)$, subject to the requirement that each optimized Tweedie prediction $\hat{z}_{0|T}^i$ remains within a bounded displacement from its reference anchor $\hat{z}_{0|T}^{i,\mathrm{ref}}$:
\begin{equation}
    \min_{Z_T}\; -\log\det(S + \epsilon I) \quad \text{s.t.} \quad \left\| \hat{z}_{0|T}^i - \hat{z}_{0|T}^{i,\mathrm{ref}} \right\|^2 \le \delta, \quad \forall\, i \in \{1,\dots,B\}.
    \label{eq:constrained_appendix}
\end{equation}
Introducing non-negative Lagrange multipliers $\{\lambda_i\}_{i=1}^B$, one per constraint, the Lagrangian is:
\begin{equation}
    \mathcal{L}_{\mathrm{gen}}(Z_T, \Lambda) \;=\; -\log\det(S + \epsilon I) \;+\; \sum_{i=1}^{B} \lambda_i \left( \left\| \hat{z}_{0|T}^i - \hat{z}_{0|T}^{i,\mathrm{ref}} \right\|^2 - \delta \right).
    \label{eq:lagrangian_appendix}
\end{equation}
Since all $B$ samples share the same text condition, backbone, and noise prior $\mathcal{N}(\mathbf{0}, I)$, the constraints are symmetric across samples.
We therefore impose a uniform multiplier $\lambda_i = \lambda$ for all $i$, giving:
\begin{equation}
    \mathcal{L}_{\mathrm{gen}}(Z_T, \lambda) \;=\; -\log\det(S + \epsilon I) \;+\; \lambda \sum_{i=1}^{B} \left\| \hat{z}_{0|T}^i - \hat{z}_{0|T}^{i,\mathrm{ref}} \right\|^2 - \lambda B\delta.
\end{equation}
Since the constant $-\lambda B\delta$ is independent of $Z_T$ and vanishes under $\nabla_{Z_T}$, we drop it to obtain the final objective in \cref{eq:loss}, where $\lambda \geq 0$ is the single hyperparameter governing the diversity--fidelity trade-off. $\square$

\begin{table}[t]
  \centering
  \caption{Gradient dynamics during DetNO optimization on SD1.5 ($B{=}5$, 20 prompts $\times$ 3 seeds). $R_z > 1$ at 8/10 steps ($^{***}p{<}0.001$, $^{**}p{<}0.01$, $^{*}p{<}0.05$).}
  \label{tab:grad_dynamics}
  \setlength{\tabcolsep}{3pt}
  \small
  \resizebox{\linewidth}{!}{
  \begin{tabular}{lcccccccccc}
  \toprule
  Metric & Step 0 & Step 1 & Step 2 & Step 3 & Step 4 & Step 5 & Step 6 & Step 7 & Step 8 & Step 9 \\
  \midrule
  \rev{$\mu_{\min}$} & $0.244$ & $0.349$ & $0.304$ & $0.256$ & $0.230$ & $0.226$ & $0.234$ & $0.248$ & $0.261$ & $0.271$ \\
  Effective rank & $2.856$ & $3.246$ & $3.065$ & $2.867$ & $2.754$ & $2.752$ & $2.802$ & $2.871$ & $2.931$ & $2.963$ \\
  $R_z$ & $1.138^{***}$ & $0.891$ & $0.975$ & $1.106^{***}$ & $1.197^{***}$ & $1.207^{***}$ & $1.173^{***}$ & $1.125^{***}$ & $1.086^{**}$ & $1.063^{*}$ \\
  \bottomrule
  \end{tabular}
  }
\end{table}

\subsection{Empirical Validation of Spectral Gradient Amplification.}
To validate that the amplification of \cref{prop:spectral} holds throughout actual
optimization, we track gradient dynamics during DetNO on SD1.5 ($B{=}5$,
$N_{\text{opt}}{=}10$, 20 MS-COCO prompts, 3 seeds each) and summarize the results in
\cref{tab:grad_dynamics}.
\rev{We deliberately extend optimization beyond the deployed $N_{\mathrm{opt}}{=}3$ so that the predicted rise and decay of the gradient ratio is traced with per-step significance; deployment corresponds to the first three steps.}
\del{$\lambda_{\min}$}\rev{$\mu_{\min}$} remains below the threshold $0.447$ across all optimization steps,
confirming that the sufficient condition of \cref{prop:spectral} is satisfied throughout.
The gradient ratio $R_z$ exceeds 1 at 8 out of 10 steps with statistical significance,
and decreases monotonically from \del{Step~3}\rev{Step~5} onward as the effective rank increases\del{ —}\rev{,}
consistent with the self-attenuating behavior predicted by \cref{prop:spectral}.
The two non-significant steps (Step~1 and Step~2) coincide with a transient rise in
\del{$\lambda_{\min}$}\rev{$\mu_{\min}$} and effective rank at the onset of optimization, after which both
quantities stabilize and $R_z$ recovers above 1 for the remainder of the trajectory.

\begin{table}[t!]
\centering
\caption{Hyperparameter configuration per backbone.}
\label{tab:hyperparams}
\begin{tabular}{lccccccc}
\toprule
Backbone & $w$ & $\eta$ & $\lambda$ & $N_{\mathrm{opt}}$ & $B$ & NFE & $\epsilon$ \\
\midrule
SD1.5 & 16 & 0.01 & 0.05 & 3 & 5 & 50 & $10^{-5}$ \\
SDXL & 16 & 0.01 & 0.05 & 3 & 3 & 50 & $10^{-5}$ \\
SD3 & 32 & 0.01 & \rev{0.01} & 3 & 5 & 28 & $10^{-5}$ \\
\bottomrule
\end{tabular}
\end{table}

\subsection{Hyperparameter Configuration}
\label{sec:hyperparams}
Tab.~\ref{tab:hyperparams} summarizes the hyperparameters used for each backbone in all
main experiments.
The pooling window $w$ is set larger for SD3 to preserve finer spatial structure,
consistent with the ablation in Fig.~\ref{fig:w_lambda_pareto}.
The fidelity weight $\lambda$ provides a transparent diversity--quality trade-off and can
be adjusted independently of the other hyperparameters; its effect on the trade-off is
reported in Fig.~\ref{fig:w_lambda_pareto}.
The batch size $B$ follows the protocol of~\cite{kim2025diverse}: $B{=}5$ for SD1.5 and SD3, and
$B{=}3$ for SDXL.
NFE denotes the number of function evaluations used by the sampler: 50 steps for SD1.5
and SDXL (DDIM), and 28 steps for SD3 (Euler), following each model's default
configuration.

\subsection{Diversity Propagation Across Timesteps}
\label{sec:tweedie_intermediate}

A natural question is whether the diversity advantage at $t{=}T$ persists
through the denoising trajectory or is gradually erased.
We measure the mean pairwise cosine similarity of the Tweedie predictions
$\hat{\mathbf{z}}_{0|t}$ at every denoising step, averaged over 20 prompts
from MS-COCO with $B{=}5$ and NFE${=}50$.
Lower cosine similarity indicates greater diversity among the $B$ predictions.

\begin{table}[t]
    \centering
    \caption{%
        Mean pairwise cosine similarity of Tweedie predictions at key timesteps.
        Averaged over 20 MS-COCO prompts ($B{=}5$, NFE${=}50$, SD1.5).
        Lower indicates more diverse predictions.
        DetNO maintains the lowest similarity throughout the denoising trajectory.
    }
    \label{tab:tweedie_intermediate}
    \begin{tabular}{lccc}
        \toprule
        Method & $t{=}T$ (step 0) & $t{=}0.5T$ (step 25) & $t{\approx}0$ (step 49) \\
        \midrule
        DDIM & $0.548 \pm 0.138$ & $0.133 \pm 0.078$ & $0.120 \pm 0.073$ \\
        CNO & $0.444 \pm 0.200$ & $0.115 \pm 0.092$ & $0.101 \pm 0.083$ \\
        DetNO & $\mathbf{0.403 \pm 0.179}$ & $\mathbf{0.032 \pm 0.057}$ & $\mathbf{0.026 \pm 0.052}$ \\
        \bottomrule
    \end{tabular}
\end{table}

\begin{table}[t]
  \centering
  \caption{Sigma ablation on SD1.5 ($B$=5, $N_{\mathrm{opt}}$=3, $\lambda$=0.05, $w$=16).}
  \label{tab:sigma_ablation}
  \resizebox{\linewidth}{!}{%
  \begin{tabular}{l|cccc|ccc|cc}
  \toprule
  $\sigma$ & Prec$\uparrow$ & Recall$\uparrow$ & Den$\uparrow$ & Cov$\uparrow$ 
           & CLIP$\uparrow$ & Pick$\uparrow$ & IR$\uparrow$ 
           & MSS$\downarrow$ & Vendi$\uparrow$ \\
  \midrule
  0.01 & \underline{0.6698} & 0.6369 & \underline{0.5200} & \underline{0.3314} & \underline{31.3793} & \underline{21.5105} & \underline{0.2444} & 0.1588 & 4.7121 \\
  0.5 & 0.6687 & 0.6389 & 0.5141 & 0.3289 & \textbf{31.3807} & \textbf{21.5129} & \textbf{0.2519} & 0.1579 & 4.7160 \\
  1.0 & 0.6554 & 0.6502 & 0.4945 & 0.3209 & 31.2677 & 21.4233 & 0.1796 & 0.1390 & 4.7669 \\
  dynamic median & 0.6538 & \underline{0.6561} & 0.5003 & 0.3214 & 31.2722 & 21.4133 & 0.1675 & \underline{0.1361} & \underline{4.7733} \\
  median (ours) & \rev{\textbf{0.6744}} & \rev{\textbf{0.6807}} & \rev{\textbf{0.5541}} & \rev{\textbf{0.3570}} & \rev{31.1396} & \rev{21.2536} & \rev{0.1088} & \rev{\textbf{0.1229}} & \rev{\textbf{4.8024}} \\
  \bottomrule
  \end{tabular}%
  }
\end{table}

\cref{tab:tweedie_intermediate} reports the cosine similarity at three representative timesteps: $t{=}T$ (initial), $t{=}0.5T$ (midpoint), and $t{\approx}0$ (final).
The results confirm that the initial diversity advantage established at $t{=}T$ is preserved throughout the denoising process. At $t{=}T$, DDIM's Tweedie predictions exhibit high cosine similarity ($0.548$), indicating mode collapse, while DetNO achieves the lowest similarity ($0.403$). This gap widens as denoising progresses: at $t{\approx}0$, DetNO's predictions are nearly orthogonal ($0.026$) compared to DDIM ($0.120$) and CNO ($0.101$).
This validates our single-step optimization strategy: diversifying the Tweedie prediction at $t{=}T$ is sufficient to produce diverse final images without per-step intervention.

\subsection{Effect of Kernel Bandwidth $\sigma$}
\label{sec:sigma_ablation}

The RBF kernel bandwidth $\sigma$ controls the sensitivity of the Gram matrix 
to inter-sample distances in the pooled Tweedie prediction space.
We compare five configurations: two fixed small values ($\sigma \in \{0.01, 0.5\}$), 
a fixed moderate value ($\sigma = 1.0$), a dynamic median updated at every 
optimization step, and our default choice of a fixed median heuristic computed 
once from the reference predictions (Eq.~\ref{eq:median}).

As shown in Tab.~\ref{tab:sigma_ablation}, small fixed values ($\sigma=0.01, 0.5$) 
yield near-uniform Gram matrices where all off-diagonal entries approach 1, 
providing weak diversity signal and resulting in the lowest VendiScore and \del{MSS}\rev{the highest MSS}.
$\sigma=1.0$ and the dynamic median improve diversity metrics but at a moderate 
quality cost. Our fixed median heuristic achieves the best MSS and VendiScore 
while also attaining the highest Precision, Density, and Coverage, confirming 
that computing $\sigma$ once from reference predictions provides a well-calibrated 
and stable operating point without requiring per-step bandwidth updates.

\begin{table}[t]
  \centering
  \caption{Computational cost comparison (SD3, NFE=28, $B=5$, 1024$\times$1024).}
  \label{tab:computational_cost}
  \resizebox{0.7\linewidth}{!}{
  \begin{tabular}{lccc}
  \toprule
  Method & Time/batch (s) $\downarrow$ & GPU Mem. (GB) $\downarrow$ & Extra Fwd. Passes \\
  \midrule
  SD3 (Baseline) & 16.3 & 28.3 & 0 \\
  CNO & 17.7 & 28.3 & $N_{\text{opt}}+1=4$ \\
  DetNO (Ours) & 17.8 & 28.3 & $N_{\text{opt}}+1=4$ \\
  \bottomrule
  \end{tabular}
  }
\end{table}

\subsection{Computational Overhead}
\cref{tab:computational_cost} compares the computational cost across methods.
Following CNO, we apply a stop-gradient operator on the UNet output when computing Tweedie predictions, which eliminates the need to retain UNet activations during backpropagation. 
As a result, DetNO achieves the same GPU memory footprint as CNO (28.3\,GB) and requires the same number of extra forward passes ($N_{\text{opt}}+1$). 
The marginal time overhead relative to CNO (17.8\,s vs.\ 17.7\,s) stems solely from the Gram matrix computation and its matrix inversion, both of which are negligible compared to the UNet forward passes.

\begin{table}[t]
\centering
\caption{Ablation: Full Gradient vs.\ Stop Gradient (SD3, MS-COCO, seed=42, single run)}
\label{tab:abl_gradient}
\resizebox{0.8\linewidth}{!}{%
\begin{tabular}{lccccccccc}
\toprule
\textbf{Method} & \textbf{Prec}$\uparrow$ & \textbf{Recall}$\uparrow$ & \textbf{Den}$\uparrow$ & \textbf{Cov}$\uparrow$ & \textbf{CLIP}$\uparrow$ & \textbf{Pick}$\uparrow$ & \textbf{IR}$\uparrow$ & \textbf{MSS}$\downarrow$ & \textbf{Vendi}$\uparrow$ \\
\midrule
Baseline & 0.6261 & \underline{0.5223} & \textbf{0.4527} & 0.2480 & \rev{\textbf{31.8805}} & \rev{\textbf{22.5719}} & \rev{\textbf{1.0503}} & \rev{0.2944} & \rev{4.2517} \\
Full Gradient & \textbf{0.6385} & \textbf{0.5309} & \underline{0.4520} & \textbf{0.2499} & \underline{31.8775} & 22.5541 & \underline{1.0468} & \textbf{0.2824} & \textbf{4.2964} \\
Stop Gradient & \underline{0.6303} & 0.5184 & 0.4437 & \underline{0.2488} & 31.8400 & \underline{22.5597} & 1.0465 & \underline{0.2858} & \underline{4.2892} \\
\bottomrule
\end{tabular}%
}
\end{table}

\subsection{Effect of Gradient Computation Strategy.} 
As shown in Table~\ref{tab:abl_gradient}, stop-gradient already yields \del{consistent improvements over the baseline across both diversity and quality metrics}\rev{clear diversity gains over the baseline (MSS, VendiScore) at a marginal quality cost}.
Full gradient achieves moderately stronger gains \del{— particularly in Recall and VendiScore —}\rev{(particularly in Recall and VendiScore)} by propagating richer second-order information through the denoising network, but at the cost of storing all intermediate UNet activations during backpropagation.
Stop-gradient closes most of this gap while reducing peak memory to the level of standard inference (\cref{tab:computational_cost}), making it the more practical default; we adopt it in Algorithm~1 and note that full gradient can be used when memory budget permits.

\subsection{Ablation on Noise Scale}
\label{sec:noise_scale}

To validate that the diversity gains of DetNO cannot be replicated by simply scaling the initial noise variance, we evaluate $\mathbf{z}_T \sim \mathcal{N}(\mathbf{0}, s^2\mathbf{I})$ across a range of scale values $s$ on SD1.5. 
As shown in Tab.~\ref{tab:ablation_noise_scale}, any deviation from the standard $s=1$ causes severe image degradation across all quality metrics. 
At $s=0.5$, Recall collapses to $0.0002$ and ImageReward drops to $-2.28$, indicating that the denoiser operates far outside its training distribution and produces degenerate outputs. Increasing the scale to $s \geq 1.2$ yields similarly catastrophic failures: Recall falls below $0.04$, CLIPScore drops by more than $8$ points, and ImageReward becomes strongly negative across all configurations. 
This behavior is expected\del{ —}\rev{:} diffusion model denoisers are trained exclusively with $s=1$, and any perturbation to the noise scale pushes the initial latent out of the distribution seen during training, causing the reverse process to produce corrupted images regardless of the direction of the scale change. 
These results confirm that noise scale manipulation is not a viable mechanism for diversity enhancement, and that the diversity gains of DetNO arise from structured optimization within the valid noise distribution rather than from distributional shift.
\begin{table}[t]                                                                                                  
  \centering                                                                                                      
  \caption{Ablation on noise scale $s$ (SD1.5). $\mathbf{z}_T \sim \mathcal{N}(\mathbf{0}, s^2 \mathbf{I})$.}     
  \label{tab:ablation_noise_scale}
  \resizebox{\textwidth}{!}{%
  \begin{tabular}{c|cccc|ccccc}                                                                                   
  \toprule
  $s$ & Prec$\uparrow$ & Recall$\uparrow$ & Density$\uparrow$ & Coverage$\uparrow$ & CLIP$\uparrow$ & Pick$\uparrow$ & IR$\uparrow$ & MSS$\downarrow$ & Vendi$\uparrow$ \\
  \midrule
  0.5 & 0.9963 & 0.0002 & \textbf{3.6769} & 0.0149 & 18.3868 & 17.7436 & -2.2765 & 0.5097 & 2.9807 \\
  0.8 & 0.7752 & 0.0194 & \underline{0.8544} & \underline{0.0688} & \underline{26.2606} & \underline{19.2687} & \underline{-1.8094} & \underline{0.2623} & \underline{4.3496} \\
  1.0 (DDIM) & 0.7005 & \textbf{0.6061} & 0.6112 & \textbf{0.3631} & \textbf{30.9096} & \textbf{21.3413} & \textbf{0.1086} & \textbf{0.1509} & \textbf{4.7352} \\
  1.2 & 0.6151 & \underline{0.0385} & 0.2208 & 0.0099 & 22.1671 & 17.3286 & -2.0955 & 0.2787 & 4.3096 \\
  1.5 & \underline{0.9984} & 0.0000 & 0.7066 & 0.0009 & 18.9369 & 16.0270 & -2.2767 & 0.5503 & 3.1409 \\
  2.0 & \textbf{1.0000} & 0.0000 & 0.7958 & 0.0009 & 19.3291 & 15.8109 & -2.2789 & 0.6051 & 2.8754 \\
  \bottomrule
  \end{tabular}%
  }
\end{table}

\begin{figure}
    \centering
    \includegraphics[width=1\linewidth]{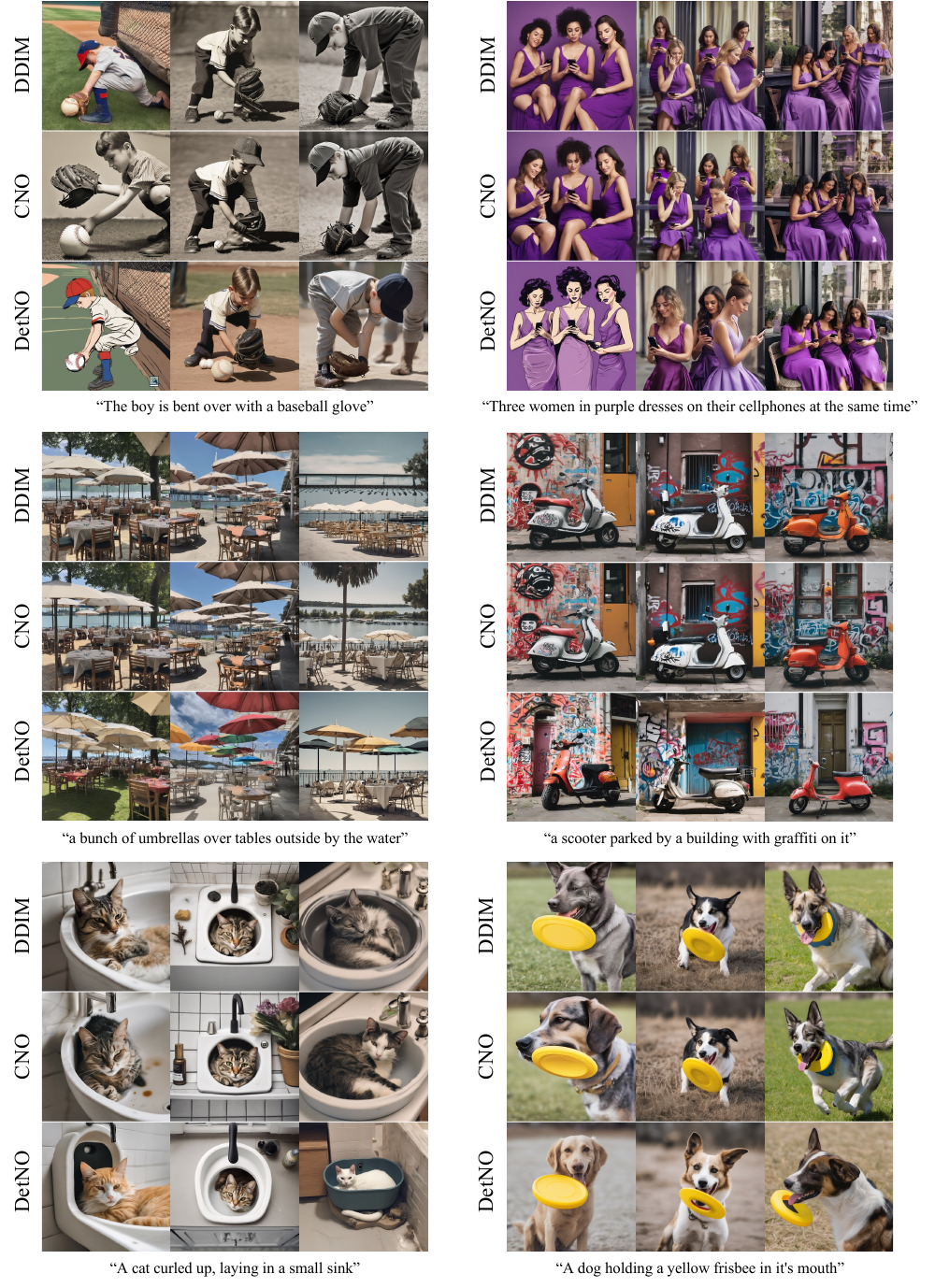}
    \caption{Qualitative results on SDXL.}
    \label{fig:supple_sdxl}
\end{figure}

\begin{figure}
    \centering
    \includegraphics[width=0.8\linewidth]{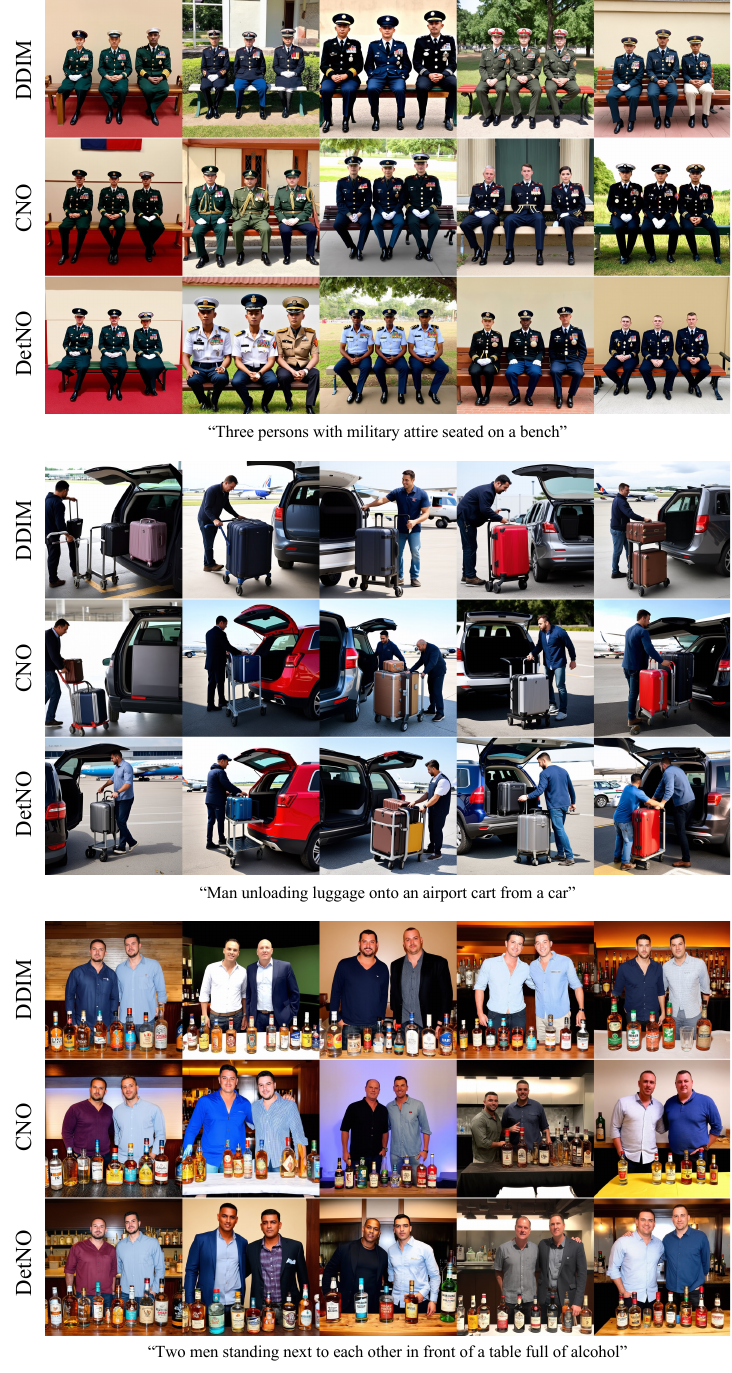}
    \caption{Qualitative results on SD3}
    \label{fig:supple_sd3}
\end{figure}

\end{document}